\documentclass{article}

\usepackage[utf8]{inputenc}%(only for the pdftex engine)
\usepackage[small]{dgruyter}
\usepackage{microtype}
\usepackage{graphicx}
\usepackage{multicol,multirow}
\usepackage{amsmath,amssymb,amsfonts}
\usepackage{rotating}
\usepackage{appendix}
\usepackage[sort&compress]{natbib}
\usepackage{hyperref}
\usepackage{xcolor}
\begin{document}

  \articletype{...}

  \author*[1]{Pablo Gamallo}
  %\author[2]{...}
  %\author[1]{...} 
  \runningauthor{Pablo Gamallo}
  \affil[1]{Centro Singular de Investigación en Tecnoloxías Intelixentes (CiTIUS), Universidade de Santiago de Compostela, Galiza/Spain.}
  %\affil[2]{...}
  %TC:ignore
  \title{Contextualized Word Senses: From Attention to Compositionality}
  \runningtitle{From Attention to Compositionality}
  %\subtitle{...}
  \abstract{The neural architectures of language models are becoming increasingly complex, especially that of Transformers, based on the attention mechanism.  Although their application to numerous natural language processing tasks has proven to be very fruitful, they continue to be models with little or no interpretability and explainability. One of the tasks for which they are best suited is the encoding of the contextual sense of words using contextualized embeddings. In this paper we  propose a transparent, interpretable, and linguistically motivated strategy for encoding the contextual sense of words by modeling semantic compositionality. Particular attention is given to dependency relations and semantic notions such as selection preferences and paradigmatic classes. A partial implementation of the proposed model is carried out and compared with Transformer-based architectures for a given semantic task, namely the similarity calculation of word senses in context. The results obtained show that it is possible to be competitive with linguistically motivated models against the black boxes underlying complex neural architectures.}
  \communicated{...}
  \dedication{...}
  \received{...}
  \accepted{...}
  \journalname{Linguistics Vanguard}
  \journalyear{2022}
  \journalvolume{}
  \journalissue{..}
  \startpage{1}
  \aop
  \DOI{...}
%TC:endignore
\maketitle

\section{Introduction}
Although the success of applying deep neural networks in natural language processing is undeniable and compelling, there are well-founded criticisms about its inability to mimic the generalization capacity, which is probably the main characteristic of human intelligence \citep{Lake2017,Marcus2018}. One of the most common and recurrent criticisms is related to the notion of compositionality as a mechanism to produce systematic generalizations.
The principle of compositionality was introduced by Frege in the early 20th century and reformulated more recently by \citet{Montague1970} and \citet{Partee2004}. It states that the meaning of a complex expression is determined by the meaning of its component parts and the way in which they are combined. Even though some kind of compositional behaviour can emerge from the outputs returned by deep neural networks, it is not clear how these systematic generalizations are carried out.

The work of \citet{Nefdt2020}, \citet{Baroni2020} and \citet{Dankers2022} suggests that large neural models are able to generalize and capture composite meaning without learning the explicit rules governing the combination of components. We are not in a position to determine whether these generalization methods are more or less efficient than those we humans use to construct meaning. They are probably different, as shown by the enormous amount of information that an artificial system requires for the properties of generalization and comprehension to emerge, with respect to the input that the human brain receives in language acquisition \citep{Warstadt2020}. \citet{Ettinger2020} argues that what probably makes the human brain and artificial neural models different is that the former extracts the meaning of sentences by compositional mechanisms to make judgments of truth, while the latter rely on the technique of prediction in context. In human acquisition, compositional abilities are at the core of the semantic interpretation process and prediction is only used to strengthen these abilities. By contrast, in artificial neural networks, prediction is the core process of self-learning and compositional abilities only emerges from massive prediction on huge amounts of textual data. In sum, predictions in large language models are at the heart of the learning process and are driven mainly by superficial contextual cues, rather than by robust and systematic representations of context meaning \citep{Pandia2021}.

%It has been \replaced{observed}{suggested} 
It has been observed that lack of compositionality is one of the main reasons why neural networks, unlike humans, rely on large amounts of data to make correct generalizations \citep{Lake2017}. However, even with those large amounts of training data, the neural networks can be easily fooled by  adversarial examples that pose no problem for humans \citep{Ebrahimi2018}.  This situation has led some researchers to emphasize data quality by proposing strategies that focus more on the quality of the input data than on the big quantity of information, which makes large language models \textit{statistical parrots} \citep{Bender2021}.
% ref:Javid Ebrahimi Daniel Lowd, and Dejing Dou. On adversarial examples for character-level neural machine translation. In Proceedings of COLING, pages 653–663, Santa Fe, New Mexico, USA, 2018}.
% Emily Bender et al 2021: On the Dangers of Stochastic Parrots: Can Language Models Be Too Big?

Given the mentioned weaknesses of neural networks, the question arises as to  whether it is possible to provide them with more modular and structured knowledge, making them more compositional, while maintaining their strengths: fast, efficient and with great adaptive capacity. In fact, interest in neuro-symbolic systems is growing up with the intention of compensating for the shortcomings of systems based purely on artificial neural models \citep{Marcus&Davis2019}. 
While the composition rules of symbolic methods are transparent and easily interpretable, they might be too rigid to cope with noise and fuzzy constructions of natural language. Neural networks, by contrast, seem much better adapted to noisy scenarios, but they are incapable of carrying out, as we have already said, systematic routines of a compositional nature essential for natural language processing \citep{Marcus2003}. The combination of both strategies, neural networks and symbolic strategies, could be a fruitful step to rethink  large language models  and make them more transparent and interpretable.

The objective of this paper is to propose a new symbolic-based language modeling strategy to encode the sense of words in context. More precisely, we define a semantic method that uses compositional rules applied on dependency trees to both build the meaning of complex expressions and to contextualize the sense of the constituent words. In addition, we compare our linguistically motivated method  with the attention mechanism of neural-based Transformers in producing contextualized vectors for each word of the input sentence.  The main difference between the two methods is that our syntax-based strategy makes use of an explicit compositional approach while the attention mechanism integrated in the neural architecture does not build contextualized senses with explicit  compositional semantic operations \citep{Wang2017}.

The proposed compositional method is not an inscrutable black box, but a robust, transparent and simple linguistic mechanism. The method is based on a compositional and incremental interpretation of syntactic dependencies, which allows to compute semantic representations of complex expressions.

In the present article, an extensive and detailed linguistic description of the method presented in previous works  \citep{GamalloICAART2021,GamalloAPPLIEDSCIENCE2021} is given. Taking into account the linguistic profile of the journal's audience, we have sought to deepen the linguistic concepts of the proposed method to the detriment of the more technical details of its implementation.
 In addition, we have included a discussion of both the principle of compositionality and the mechanism of attention to make it clearer the conceptual difference between the proposed compositional method and Transformers. Finally, we have grouped together in one section all the experiments that had been carried out in previous works.

%Organization of the paper: related work, compositional strategy, experiments, conclusions
This article is organized as follows. In the next section (\ref{sec:relatedwork}) we explore how both neural networks and symbolic strategies deal with compositional meaning and word contextualization. Then, in Section \ref{sec:method}, our dependency-based compositional model is described. Section \ref{sec:experiments} reports a semantic experiment aimed at comparing the contextualized vectors produced by our strategy with those generated by Transformer-based large models. Finally, conclusions and future work are addressed in Section \ref{sec:conclusions}.  
%%%%%
%(Paradox of compositionality, Dankers 2022) 
%-Partee, B. (2004). Compositionality in formal semantics. Oxford: Blackwell.
%``The meaning of a complex syntactic expression is a function of the meanings of itsmmediate constituents and the syntactic rule used to combine these constituents.'' (Nefdt 2020)
%-Brenden Lake, Tomer Ullman, Joshua Tenenbaum, and Samuel Gershman. Building machines that learn and think like people. Behavorial and Brain Sciences, 40:1–72, 2017. 
%-Gary Marcus. Deep learning: A critical appraisal. https://arxiv.org/abs/1801.00631, 2018.

%(Baroni) what causes the surprising and sometimes dramatic failures that still affect Neural Networks [8, 9]:
%(Wan 2017): , and the entire networks are optimised for the classification task using supervised training over relatively smaller datasets.

\section{Neural and Symbolic Based Strategies}\label{sec:relatedwork}

%%objective of the paper: contextualized words in a symbolic way (could be neuro-symbolic...)....
%MY PROPOSAL-> we propose a linguistically structured architecture that is able to capture semantic compositionality while keeping similar efficiency and robustness to neural networks.
The meaning of a complex expression relies on the contextualized senses of its constituent words. To build the meaning of complex expressions and the senses of their constituent words as contextualized vectors, two methods can be followed:

\begin{itemize}
\item Methods (mostly neural-based) that combine word vectors without making use of explicit compositional operations.
\item Symbolic-based methods that rely on explicit compositional rules linguistically motivated  by syntactic-semantic functions.
\end{itemize}

In both strategies, distributional representations encoded in continuous vectors are a proxy for meaning.

\subsection{Neural-Based Architectures}

In the first type of methods, no explicit linguistic information or no explicit notion of linguistic trees is required. The basic approach (not strictly neural) is to combine embeddings of two co-occurring words with arithmetic operations: addition or component-wise multiplication \citep{Mitchell&Lapata2008,Mitchell&Lapata2009,Mitchell&Lapata2010}. This approach is not compositional because the syntactic functions that links  words is not considered, and so two sentences with the same constituents but with different functions are represented in the same way (e.g. \emph{``Federer beat Nadal''} and \emph{``Nadal beat Federer''}). Much more sophisticated and complex approaches are found in neural networks, for instance in the use of recurrent neural networks such as Long Short Term Models (LSTM) to process sequentially the input sentences, or more recently networks based exclusively on attention mechanisms \citep{Vaswani2017}, called Transformers. Attention mechanisms replace sequential processing, which is difficult to parallelize, with highly efficient and parallelizable distributed processing,  designed to account for long-distance word relationships.

\begin{figure}
  \includegraphics[scale=0.30]{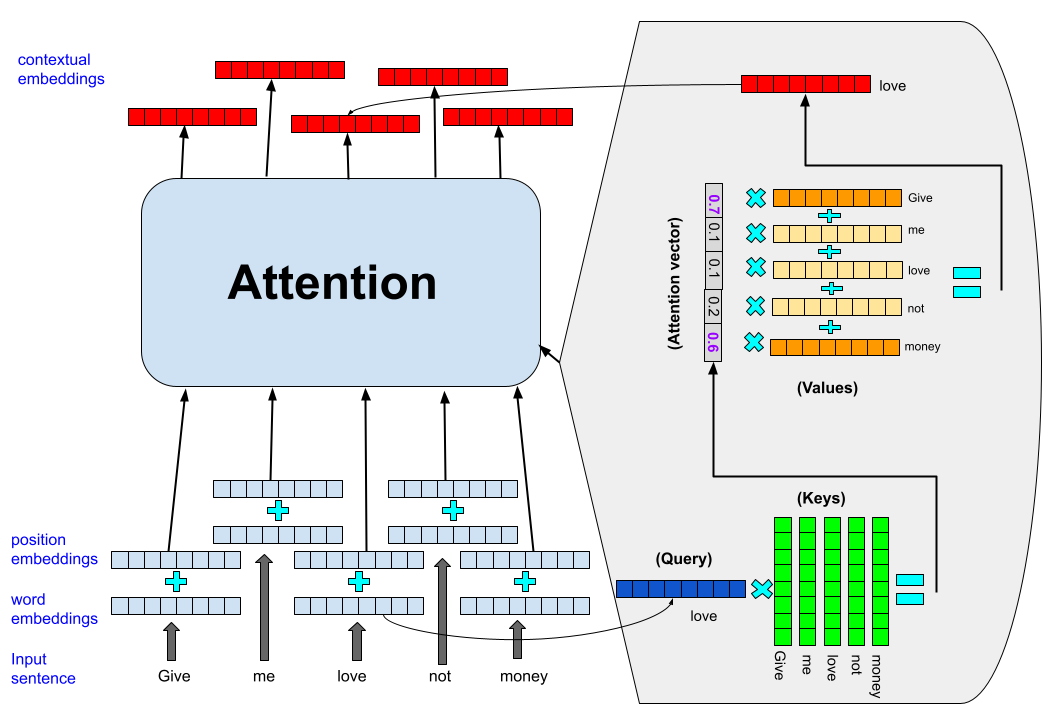}
 \caption{Simplified architecture of the self-attention mechanism} \label{fig:attention}
\end{figure} 

Figure~\ref{fig:attention} shows a simplified version of the self-attention mechanism underlying the Transformer architecture, one of the most recent and powerful neural-based models. On the left, a sentence is processed by associating to each constituent word a static word embedding, which was pre-trained from a corpus. Each word is assigned a static embedding out of context. This non contextual embedding is added up to a positional embedding that has been built by considering the position of each word in the sentence. The resulting word embeddings (static + position) for all words of the sentence represent the input of the attention mechanism, which returns a contextualized vector for each word as a result.

On the right of Figure~\ref{fig:attention}, we show how the contextualized embedding is built in the attention module for a specific word (\emph{love}). The attention strategy is inspired from the information retrieval approach as it evokes the key, value and query concepts. A vector conceived as the query is compared to a list of other vectors (the keys), so as to find the best candidates (values) for the initial query. This allows the module to focus on a word and look at other words (keys) in the input sentence so as to grasp their relevance with respect to the given word (query). The objective is to calculate the degree of association (or attention) between each word and the rest of the sentence by considering their position. In the attention mechanism, this idea is implemented as follows. Let us start with the sentence \textit{``Give me love not money''} and the word \textit{love}: the initial static embedding of \textit{love} (along with its positional embedding) is projected into three different vectors: query, key and value vectors. The same is done for the rest of words. These three vectors are created by multiplying the initial word embedding by three matrices with weights adjusted during the training phase. Then, the query vector of \textit{love} is compared against the key vectors associated to all the words in the sentence by computing the dot product. The dot product of \textit{love} with a key vector gives rise to an attention score. This is done with all words in the sentence resulting in an attention vector of 5 dimensions, one for each word. The higher the attention score, the closer the linguistic proximity in the sentence between the query and the key. In the specific toy example of   Figure~\ref{fig:attention}, the highest scores (represented by a lighter color) correspond to the key vectors of \textit{Give} and \textit{money}, given that \textit{love} is the direct object of the verb \textit{Give} and it plays the same semantic role as \textit{money} in the sentence. Ideally, each word tends to "attend" to those with which it would be linguistically linked by means of syntactic, semantic, discursive, or pragmatic relationships.

The final step of the attention mechanism, represented in the upper right part of the figure, consists of adding up all the value vectors of the 5 words by taking into account the weight of the attention score of each word with respect to \textit{love}. It results in the contextualized vector of \textit{love}. The same is done for the rest of the words in the sentence. The whole process is called \textit{self-attention}. The prefix 'self' is used as the attention is directed to words of the same language. In a translation context with an encoder-decoder architecture, \textit{cross-attention} is carried out.

Yet, there are more complex attention architectures. In a multi-head attention architecture, self-attention is repeated and performed $N$ times in parallel with different query/key/value vectors for each work of the input sentence.  In addition, the whole multi-head attention process is replicated for each attention layer of the neural network, which are connected by feed forward layers.

Considering that in a single self-attention layer an input word is assigned 6 different vectors or embeddings (static, positional+static, query, key, value and contextualized), in a multi-head architecture of 16 heads with 24 layers as in BERT-large \citep{Devlin2019}, the number of vectors required to fully encode a word in a particular position of a sentence is 1,178: 2 input vectors corresponding to the input embedding and its addition with the positional one, 384 query vectors (16 per head and 24 per layer), 384 key vectors, 384 value vectors, and 24 contextualized outputs (one per layer).  So, the total number of vectors required to encode the five words of the sentence in  Figure~\ref{fig:attention} is 5,890. Each embedding is supposed to be a linguistic representation of a word, but there are no clear linguistic or neuro-linguistic clues to interpret what kind of information is being encoded in each of these many word embeddings. The whole architecture is a black box made of a continuous adjustment of weights and values (through million or even billion parameters) which are calculated using the brute force of high-performance computing through backpropagation. In linguistic terms, we know that the attention mechanism constructs contextualized embeddings by summing the value vectors of co-present words in a sequence, giving more weight to some than to others. It is therefore interested in syntagmatic relations. In this process of contextualization, words in exclusive paradigmatic opposition do not intervene directly. So, the attention mechanism does not work directly with lexical selection preferences.

%In terms of explainability, there are interactive tools available to visualize attention in Transformer-based language models \citep{Vig2019}. These tools allow to view in which layer and on which words the attention is pointing out given a specific entry. However, this does not explain why and how the system is able to learn such systematic word relationships without any explicit linguistic knowledge. 

Even though the architecture of the attention mechanism (and also that of recurrent neural networks) seem not be motivated by explicit linguistic knowledge, the neural networks are more successful than the symbolic models on many Natural Language Processing (NLP) tasks, in part because of their ability to process very large amounts of data.

\subsection{Symbolic and Compositional Approaches}
The second type of methods to process the meaning of words in context relies on explicit compositional functions. This gives rise to symbolic architectures that are compositional by design \citep{Hupkes2019} and are often referred to as Compositional Distributional Semantics. 

The most popular approaches in Compositional Distributional Semantics design compositional models on the basis of the combinatorial behavior inspired by Categorial Grammar \citep{Steedman1996}. In these approaches, functional words (verbs, adjectives and adverbs) are represented as high-dimensional tensors that are applied on word arguments, represented as simple vectors, to modify and specify their meaning \citep{Coecke2010,Baroni2010,Grefenstette&Pulman2011,Krishnamurthy2013,Kartsaklis2013,Baroni2013,Baroni2014,Wijnholds2020}.  Two problems arise with this type of models: one of efficiency and the other rather conceptual.  On the one hand, these models result in an information scalability problem, since tensor representations grow exponentially  \citep{Kartsaklis2014}. 
%In \citet{Paperno2014}, the authors try to partially overcome the scalability problem of tensors by proposing a method, called \emph{practical lexical function model}, where a functional word is represented as a vector plus an ordered set of matrices, with one matrix for each argument the function takes. 
 On the other hand, it is not possible to assign a contextualized meaning to all words, since some are functions that contextualize others. This contradicts the co-compositional hypothesis \citep{Pustejovsky1995}, which states that two related words influence and constrain each other in the meaning construction process and, thereby, they can behave as both functions and arguments at the same time \citep{GamalloCLLT2017,GamalloJLM2019}.

Other approaches to Compositional Distributional Semantics, based on Dependency Grammar (instead of Categorial Grammar) do not make use of  n-order tensors to represent functional words.  The work by \citet{Erk&Pado2008} proposes a strategy in which the static meanings of two related words are combined by means of a syntactic dependency to give rise to two contextualized senses, one per word. The main operation in this dependency-based model lies in the selection preferences that each word imposes on the other. This strategy follows the co-compositional hypothesis stated above. The problem with this approach based on selection preferences is that it cannot be easily adapted to the interpretation of sentences of any type and size. As we will show in the next section, the main contribution of our work is, precisely, to generalize the Erk and Padó's model in order to be able to encode larger expressions such as sentences of any size. 
 Similarly, in Type Composition Logic \citep{Asher2016} the construction of the compositional meaning relies on two lexical functions that shift the meaning
(and in some coercitive cases, the semantic type too) of the two input words.  However, this approach only applies to two-word composition, and is not easily adapted to the modeling of longer expression. In \citet{Weir2016}, a dependency-based graph  captures the full sentential context of the word, but with very sparse word representations.

 The linguistic-based works we have introduced have something in common: they make use of distributional applications, including word vectorization and massive learning of lexical semantics from large corpora, to help validate linguistic hypotheses and formal models of language. 

With this same focus, \citet{Boleda2013} take as their starting point a formal modeling of the semantics of adjectives and design experiments with distributional models to check hypothesis about some semantic types. The experiments let them conclude that there is no difference between non-intensional subsective adjectives and intensional (non-subsective) ones. In a more theoretical work, \citet{McNally2014} argues that distributional representations have potential to serve as models for the semantics of very complex phenomena, such as the theory of kinds to treat generic uses of nominals.  
 It is also worth mentioning the approach called Functional Distributional Semantics 
 \citep{Emerson&Copestake2016}, which embeds  distributional information with model-theoretic semantics.

Our work, like all of the aforementioned,  make use of empirical distributions learnt from corpora to either validate or not a semantic hypothesis and contribute to the design of computational architectures that better model the semantic interpretation process in a more transparent way. By contrast, other linguistic-based approaches thoroughly explore the results of the neural models to find some glimmer of implicit linguistic knowledge in their black boxes. This task is carried out by defining specific tests to analyze whether neural-based language models can learn syntactic regularities and are provided with compositional abilities \citep{Linzen2018,Kim2020,DeDios2022}. \citet{Linzen2018b} states that the role of linguists would be to clearly delineate the linguistic capabilities that can be expected of large language models, by constructing controlled experimental tests that can determine whether those desiderata have been met. This new paradigm of linguistic research is of great interest and can be seen as very different but complementary to the one we follow.
%We consider that is a very necessary and promising approach, but it is in the opposite direction from ours, as we use distributions as an interpretative semantic space to both better understand linguistic hypothesis and figure out new computational implementations that might be different from the current architectures. 
 
 In the present work, we will implement a distributional architecture to check whether selectional preferences can dynamically build the semantic meaning of composite expressions.

\section{A Compositional Strategy Based on Syntactic Dependencies and Selection Preferences}\label{sec:method}
The objective of this section is to introduce a compositional strategy based on the notion of selection preferences in a co-compositional scenario. 

\subsection{Selectional Preferences and Paradigmatic Classes}
Given two syntactically related words, each one restricts the meaning of the other through the mutual imposition of selection preferences
  \citep{Erk&Pado2008,GamalloCLLT2017}.
 
 More precisely, the combination of two words, $w_1$ and $w_2$, related by means of the syntactic dependency $R$, gives rise to two restricted and contextualized word senses, $w_1'$ and $w_2'$, in relation $R$, where  $w_1'$  is obtained by combining the meaning of $w_1$ with the selectional preferences imposed by $w_2$ in $R$, and  $w_2'$ is the result of combining the meaning of $w_2$ with the selectional preferences imposed by $w_1$ in $R$. Selectional preferences are represented as the generic sense associated with the paradigmatic class of most relevant words appearing in one of the two syntactic roles, either \textit{head} or \textit{dependent},  of the dependency relation $R$. 

\begin{center}
\begin{figure}[h]
  \includegraphics[scale=0.35]{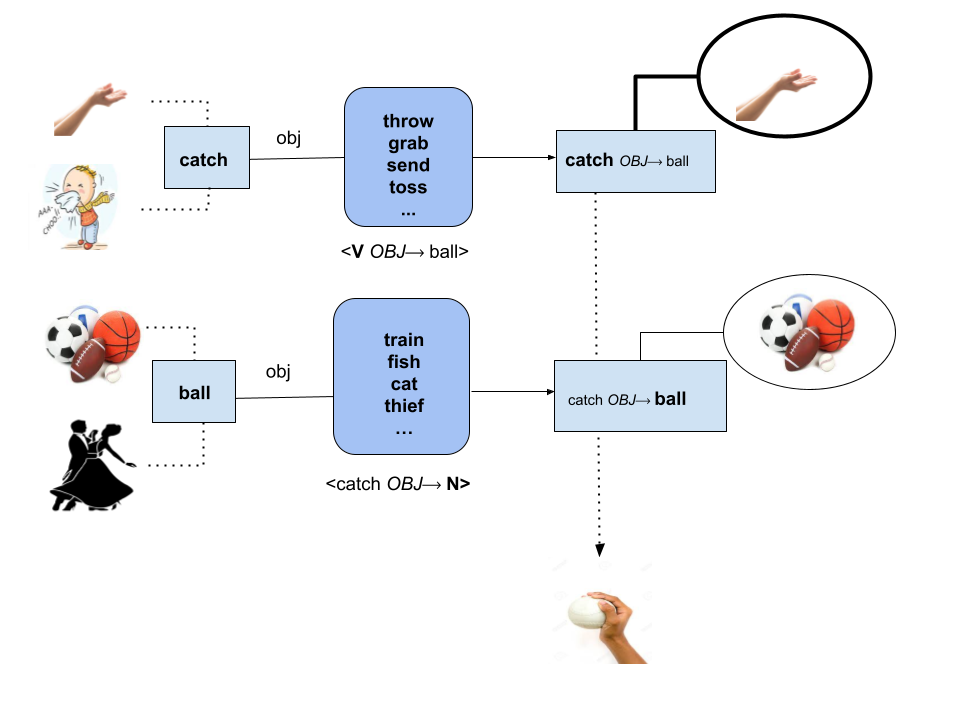}
 \caption{Co-compositional construction of the contextualized senses of \textit{catch} and  \textit{ball} by using their selectional preferences.} \label{fig:preferences}
\end{figure} 
\end{center}

Figure~\ref{fig:preferences} provides an intuitive example of this semantic procedure showing how the meaning of two polysemous words is restricted when they are combined in a particular syntactic dependency. This is the case of  the verb \textit{catch}, which can refer to either a grabbing action (represented by means of two open hands in Figure~\ref{fig:preferences}) or to the result of contracting a disease (the drawing of the doll with a cold), combined with  the noun \textit{ball}, referring to either a physical spherical object or a dancing event. 
Given the combination of the verb and the noun in the direct object relation (obj), two different compositional operations are carried out (the order is not relevant). First, in the head position, noted $<\boldsymbol{V} \,\,\, OBJ\!\!\!\rightarrow ball>$, of the direct object relation, the meaning of the verb \textit{catch} is combined with the meaning of the paradigmatic class of those most frequent verbs co-occurring with \textit{ball} in that syntactic position, namely \{throw, grasp, send, toss,\ldots\}. The meaning of the paradigmatic class, which can be seen as the selectional preferences imposed by \textit{ball}, restricts the meaning of the verb by selecting the grabbing sense.
Second, in the dependent position, $<catch \,\,\, OBJ\!\!\!\rightarrow  \boldsymbol{N}>$, the meaning of the noun \textit{ball}  is combined with the meaning of the paradigmatic class of those most frequent nouns co-occurring with \textit{catch} in that syntactic position, namely \{train, fish, cat, thief, \ldots\}. The meaning of the paradigmatic class, which represents the selectional preferences imposed by \textit{catch}, restricts the meaning of the noun by selecting the sense referring to the physical spherical object. In sum, the two words restrict, contextualize and disambiguate each other when they are combined by means of a specific syntactic dependence.

The meaning of the complex expression would be the result of combining the two contextualized senses (the picture below in the figure), but, given that Dependency Grammar does not consider categories for complex expressions, we do not address the combination between contextualized senses in the present work. We just propose that the most representative meaning of the complex expression corresponds to the contextualized sense of the root, which is, in the current example, the sense of the verb head:  $<\boldsymbol{catch} \,\,\, OBJ\!\!\!\rightarrow ball>$.
%no meaning of a complex expression because 1) in dependency grammar there is only two types of syntactic objects: words and dependencies (no complex expressions); and 2) the vector space (semantic space) is partitioned in sub-spaces associated to lexical categories (Part-of-Speech): noun, verb, adjective, and adverb vector spaces. 

%more formal definition of selection preferences
In the semantic space, out-of-context lexical words are represented as static vectors, while selection preferences are dynamic vectors resulting from adding the static vectors of the words belonging to a paradigmatic class in a syntactic position.
 In the syntactic dependency  $<w1 \,\,\, R\!\!\!\rightarrow  w2>$, the selection preferences imposed on word $w2$ by $w1$ are computed by adding the vectors of those words belonging to the paradigmatic class $\mathrm{C1}_{<w1 \,\,\, R\!\rightarrow \, C1>}$, such that $\{w \arrowvert w \in \mathrm{C1}_{<w1 \,\,\, R\!\rightarrow  \, C1>}\}$ is the set of words that can replace $w_2$ in relation $R$ with $w_1$. The contextualized sense of $w2$ is the result of combining (by component-wise multiplication or addition) the vector of $w2$ with the vector constructed with the words belonging to the class  $\mathrm{C1}_{<w1 \,\,\, R\!\rightarrow \, C1>}$. The contextualized sense of $w1$, on the other hand, results from combining the vector of $w1$ with the one built with the words belonging to the class  $\mathrm{C2}_{<C2 \,\,\, R\!\rightarrow \, w2>}$, such that $\{w \arrowvert w \in \mathrm{C2}_{<C2 \,\,\, R\!\rightarrow  \, w1>}\}$ is the set of words that can replace $w_1$ in relation $R$ with $w_2$. 
 More formally, let $\vec{w1}$ and $\vec{w2}$ be the static vectors of the two words in dependency $<w1 \,\,\, R\!\!\!\rightarrow  w2>$; then, the two contextualized senses, $\vec{w}_{<w1 \,\,\, R\!\rightarrow \, C1>}$ and $\vec{w}_{<C2 \,\,\, R\!\rightarrow \, w2>}$ ,  are two dynamic vectors computed as follows:

\begin{alignat}{2}
&   \vec{w}_{<w1 \,\,\, R\!\rightarrow \, C1>} = \vec{w1} \odot  \vec{C1} \label{eq:comp1} \\
&    \vec{w}_{<C2 \,\,\, R\!\rightarrow \, w2>}  = \vec{w2} \odot  \vec{C2} \label{eq:comp2}
\end{alignat}
where $\vec{C1}$ and $\vec{C2}$ are the vectors representing the selection preferences of the corresponding paradigmatic classes C1 and C2, and which are the result of the following operations:

\begin{alignat}{2}
&   \vec{C1} =  \sum \limits_{w \in \mathrm{C1}_{<w1 \,\,\, R\!\rightarrow  \, C1>} }  \vec{w} \label{eq:head} \\
&   \vec{C2} =  \sum \limits_{w \in \mathrm{C2}_{<C2 \,\,\, R\!\rightarrow  \, w2>}  } \vec{w} \label{eq:dep}
\end{alignat}

 It means that selection preferences are constructed by adding the word vectors belonging to a paradigmatic class. It is not necessary to consider all the words of the paradigm to get acceptable results, but just the most relevant. %The best results were obtained by selecting the $N$ most relevant words ($N=100$) of a paradigmatic class, where the relevance of a word results of computing the lexical association (i.e.,  loglikelihood or pointwise mutual information) between that word and the lexical-syntactic context characterizing the paradigm.
 
 Paradigmatic classes are constructed by massive search on large parsed corpora and static vectors are initialized as word embeddings pre-trained in tagged corpora. These embeddings are built in separated semantic spaces representing lexical categories, i.e.,  nouns, verbs, adjectives, and adverbs are separated in different vector spaces. We follow one of the fundamental principles of Cognitive Grammar, which states that syntactic categories construct different conceptualizations \citep{Langacker1987,Langacker1991} and so they project different semantic representations, even if they can evoke the same entities. 
 For instance, the meaning of the word \textit{catch} would be found in different semantic spaces depending on whether it is a verb or a noun. In our model,  the embeddings of \textit{catch} as a verb and as a noun are in different vector spaces and therefore cannot be put in the same paradigmatic class. We conceive paradigmatic classes as semantic categories with the same conceptualization and thereby constituted by words sharing the same part-of-speech. The use of paradigmatic classes allows us to align vector spaces, i.e., to make them compatible. In the expression \textit{catch the ball}, the vector of \textit{ball} is not directly combined with the vector of the verb \textit{catch} because the two vectors are made up of incompatible syntactic contexts. Nouns appear in syntactic positions different from those of verbs. However, thanks to the use of paradigmatic classes, we can combine \textit{ball} with the vectors of all nouns (or the most frequent/relevant ones) appearing as direct object of \textit{catch}, and vice-versa, the vector of \textit{catch} is combined with the vectors of all verbs having \textit{ball} as direct object. This is a fundamental difference with regard to the attention mechanism in Transformers, where attention represents syntagmatic relationships and all word embeddings are in the same vector space.

%of those verbs (\emph{eat}, \emph{jump}, etc) that are combined with the noun \emph{horse} in that syntactic context. Component-wise addition of vectors has an union effect. In more intuitive terms,  $\vec{horse}^{\circ}$  stands for the  inverse selectional preferences imposed by \emph{horse} on any verb at the subject position. As this new vector consists of verbal contexts, it lives in the same vector space as verbs and, therefore, it can be combined with the direct vector of \emph{run}.

%--start by briefly describing the concept of selection preferences in Erk and Pado work (usar figura...). This idea was implemented in \cite{GamalloJLM2019} and now we generalize the concept of selection preferences so as to deal with sentences on any size and syntactic structure. 

\subsection{Selectional Preferences in a Dependency Tree}

\begin{center} 
\begin{figure}[h]
  \includegraphics[scale=0.35]{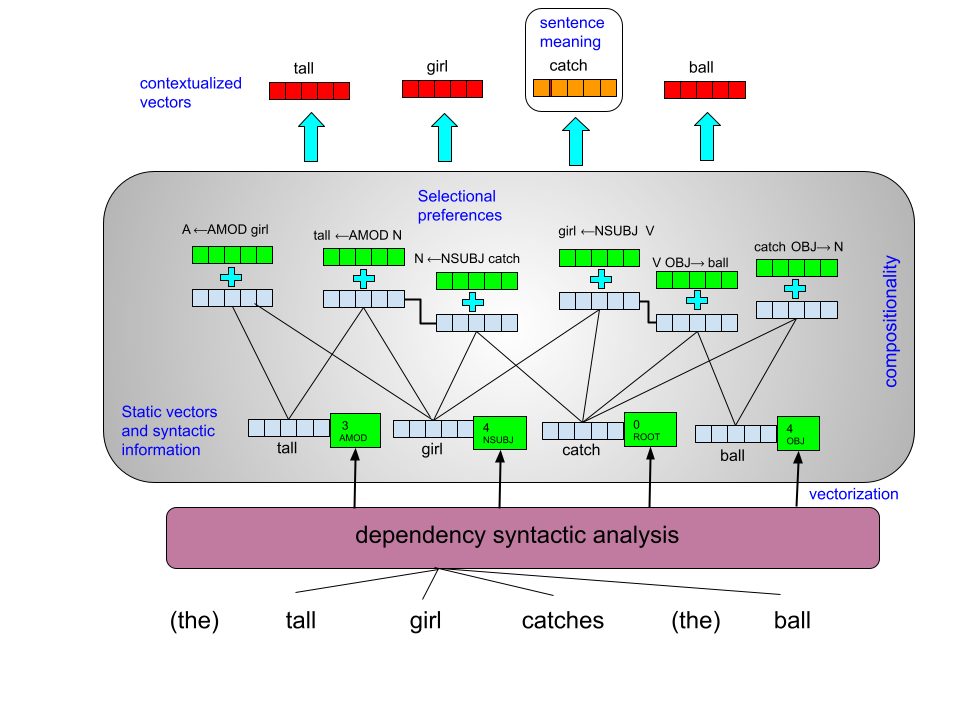}
 \caption{Simplified architecture of the dependency-based compositional approach to build contextualized word vectors.} \label{fig:architecture}
\end{figure}
\end{center}

In order to build the contextualized senses of words  in an incremental way, it is necessary to allow selectional preferences to be defined in larger dependency trees representing the syntactic analysis of sentences.

Figure~\ref{fig:architecture} shows the architecture of the dependency-based compositional approach. The input sentence is analyzed in dependencies and lexical words are initialized with static vectors defined in semantic spaces according with their lexical category. In the internal layer, selection preferences are constructed and combined with the static vectors giving rise to several contextualizations according to the number of dependencies the words are involved. The intermediate vectors associated with \textit{girl} and \textit{catch} in the figure illustrate these internal operations. Finally, the resulting contextualized vectors associated with the lexical words are the output of the compositional mechanism. Given that the verb \textit{catch} is the root of the input expression, it can be used to represent the meaning of the sentence.

%It is important to note that the proposed model is easily adaptable to usage-based syntactic-semantic theories, such as Construction Grammar (CxG) \citep{Goldberg1995}.
%CxG is a bottom-up strategy that leads to the emergence of linguistic units, called constructions,  ranging from very schematic syntactic patterns to lexical level collocations. Given this definition, we argue that a construction in CxG could be defined as a sequence of paradigmatic classes (or GSPs) represented as slot-constraints at different levels of generalization. %Taking into account grammatical cues and predefined linguistic configurations, we can redefine  GSPs at several semantic levels.   
% In this way, for instance, it is possible to define the syntactic class of nouns playing the role of subject in a bi-transitive construction without specifying the lexical units of any constituent. This gives rise to a contextualized vector representing the very generic meaning of the subject position in a bi-transitive construction. On the other hand, and in a complementary way, our approach allows us to define fully lexicalized GSPs. 
%Such a semantic flexibility underlying the proposed model  will not be explored in the experiments described in the next section, but it will be considered in further implementations. 

%\begin{mdframed}[hidealllines=true,backgroundcolor=blue!20]
  \section{Experiments} \label{sec:experiments}

  On the basis of some of the main ideas depicted in the previous section, we implemented a compositional dependency-based system,
  %\emph{DepFunc},
   \emph{DepFunc},\footnote{Freely available at 
  \url{https://github.com/gamallo/DepFunc}}
   which has been compared to several BERT-like architectures for the task of sentence similarity in four languages: English, Portuguese, Spanish and Galician. 
   Contextualized embeddings derived from autoencoder language models, such as BERT, outperform in many tasks contextualized embeddings derived from generative autoregressive models (e.g., GPT architecture) \citep{Lenci2022}.
  The experiments performed and the results obtained have already been described in detail and published  recently %\citep{Author2020,Author2021}. %In the present article, we summarize all the experiments carried out in a single table. 
\citep{GamalloICAART2021,GamalloAPPLIEDSCIENCE2021,GamalloSEPLN2022}.
 %------>DESCOMENTAR: We put together in one table the most relevant results of those experiments to show the potential of the proposed compositional method.
 For more details  (configuration of the systems, elaboration of the datasets in each language, results of all the configurations of the evaluated systems, etc.), please refer to the cited articles.

\begin{center} 
\begin{table}[ht!]
\begin{tabular}{lccccc}
\hline
\textbf{\emph{Models}} & \textbf{$\rho$(en)}  & \textbf{$\rho$(pt)} &  \textbf{$\rho$(es)}  &  \textbf{$\rho$(gl)} & \textbf{av.} \\ \hline\hline
Baseline (static)  & 29 & 28 & 29 & 28  & 29                \\ %%equivalent to vector addition as subject and object are the same words in the two compared sentences. 
\hline
DepFunc (compositional) & \textbf{53} &  \textbf{55} & \textbf{45} & 47 & 50          \\
BERT (attention)  & 45 & 43  & 42 & \textbf{57}  & 47 \\
\hline
\hline
%\text{\citep{Grefenstette&Sadrzadeh2011b}} & 28 & - & - & - & -  \\
%\text{\citep{Hashimoto2014}} & 42 & - & - & - & -  \\ 
%\text{\citep{Polajnar2015}} & 33 & - & - & - & - %\\
%\text{\citep{Wijnholds2020}} & 54 & - & - & - & - \\
%SBERT  & 61 & - & - & - & - \\
%\hline
\end{tabular}
\caption{Spearman correlation between the compared models (DepFunc and the best configurations of BERT) and human judgments on subject-verb-object sentence pairs. The table also shows a baseline based on just comparing static vectors (first row).} \label{tab:results}
\end{table}
\end{center}

Table~\ref{tab:results} compares the results obtained by our compositional strategy (DepFunc) with those returned by the best configurations of the BERT model for each language.

Concerning the results shown in Table~\ref{tab:results}, there are no major differences between the four languages, which range from 42 to 57 Spearman correlations. Nor is there much difference between the two types of models used (compositional and attention), since the compositional one achieves a macro-average score of 50 and the attention mechanism of 47 (last column in Table~\ref{tab:results}). 
%The former performs better in the English, Portuguese and Spanish dataset, while the latter performs better with the Galician test.  
This proves that the compositional model performs very competitively with respect to the Transformer-based strategy, by using a more transparent linguistic architecture with much less parameters and trained with a smaller corpus. Even though BERT and DepFunc returned similar scores, we applied the Student's test to compare their score distributions, and the results indicated that the difference between both systems is statistically significant, with p-value $\le 0.005$.

The English dataset was used for the first time in \citet{Grefenstette&Sadrzadeh2011a}. The datasets for the other three languages were built on the basis on the English one and thereby follow its structure. All are freely available and can be found in the DepFunc software. Each test dataset consists of $\approx$200  pairs of subject-verb-object triples as those in Table~\ref{tab:dataset}, along with a similarity score, in the third column, assigned by a human evaluator (or rather by the average of several ones). The scores range from 1 to 7 depending on the degree of similarity and semantic acceptability. Each pair is also assigned a Cosine similarity on the basis of the system prediction: DepFunc in column 4 and BERT in column 5, after having normalizing the scores for the range 1-7. The evaluation procedure consists of comparing the Spearman correlation between the Cosine similarity (system prediction)  and the similarity scores given by human evaluators.  Since Transformer-based models take as input sentences with determiners and inflected words, the triples of the dataset were expanded into grammatical sentences.

\begin{table}[ht!]
\begin{tabular}{lcp{1cm}cc}
\hline
\bf expr1 (in lemmas) & \bf expr2 (in lemmas) & \bf human score & \bf DepFunc & \bf BERT  \\
employ buy property & employ purchase property & 7.0 & 7.0 & 6.3  \\
firm buy politician &  firm purchase politician & 4.3 & 6.3 & 6.8\\
system provide facility & system supply facility & 6.5 & 4.3 & 7.0  \\
mother provide baby &  mother supply baby & 5.1 & 3.6 & 5.5\\
report draw attention & report attract attention & 6.5 & 4.6 & 6.9 \\
child draw picture & child attract picture & 1.5 & 3.7 & 6.0 \\
boy meet girl & boy visit girl & 4.5 & 3.9 & 6.8 \\
system meet requirement & system visit requirement & 1.5 & 2.5 & 6.5 \\
people run round & people move round & 3.3  & 3.9 & 6.8  \\
machine run application & machine move application & 1.6 & 0.0 & 6.7 \\
\hline
\end{tabular}
\caption{Sample of the English dataset. Each pair is assigned a human score, along with the similarity scores provided by both DepFunc and BERT.}\label{tab:dataset}
\end{table}

In the small sample of Table~\ref{tab:dataset}, it can be observed that the scores of BERT tend to have a reduced range to high values: between 6 and 7. On the contrary, the values of DepFunc range between 0 and 7, a range that is more similar to that used by human annotators. However, the score 0 returned by DepFunc reveals a problem of the model. This value appears when no member of the paradigmatic classes on the basis of which the selection preferences are constructed could be found in the parsed corpus. %In total, 135 pairs were assigned score 0  by DepFunc.  %This tends to occur when the analyzed expressions are provided with odd semantics and unnatural selection preferences, as the second expression ("machine move application") of the last pair in Table~\ref{tab:dataset}.

%\added{After analyzing the errors of all systems, we found that Transformed-basded models tend to overestimate similarity of the sentences since their errors correlated with those sentence pairs that are not similar. Conversely, compositional-based models have a more scattered and less biased error distribution, that is, they get confused without a clear error pattern.}

%sweat@run@face sweat@operate@face 1.15384615384615 0.435646
%employee@buy@property employee@bribe@property 1.15384615384615 0.338889
%system@run@application system@move@application 1.07692307692308 0.4357415

%We also show at the bottom of Table~\ref{tab:results} the results obtained by other systems applied to the same English dataset, although these experiments are not completely comparable to ours. In the last line, the table shows the score obtained with Sentence-BERT (SBERT), a supervised system fine-tuned with large annotated datasets for sentence entailment in English \citep{Reimers2019}. 

%And finally, it should be emphasized that there is still, in this task, a considerable difference between the system predictions and human observations, as reflected by the medium-low correlations obtained in the experiments. This indicates that better models are needed, perhaps not larger or computationally deeper, but of higher quality and with deeper linguistic knowledge.

\section{Conclusions} \label{sec:conclusions}
%%retomar o puzzle da intro: emerging compositionality but without explicit rules...
%%...
%%
We have highlighted in this work the lack of transparency and explainability of artificial neural architectures, specifically, those based on the attention mechanism. They are black boxes that manage to emerge generalizations and apparently compositional behaviors, without explicitly encoding any knowledge about the principle of compositionality.

In contrast to these large models, we have described an alternative semantic strategy based on syntactic dependencies and how these are combined to construct contextual meaning. In this paper, we have focused on how contextual vectors representing the meaning of words in an input sentence are elaborated. According to our proposal, the interpretation of a sentence is the process of building the sense of each constituent word in a recursive and incremental way,  where the contextualized sense of the root word is the most representative of the whole sentence. The proposed strategy is compositional, linguistically interpretable and does not require large computational architectures. It is based on the vectorization of paradigmatic classes, in contrast to the vectorization of words connected in a sequence by syntagmatic relations, as the attention mechanism does. In the evaluated task, the results of our strategy are competitive with those obtained with neural architectures based on the attention mechanism,  even if our models were trained with smaller corpora.

In future work, we will do a full implementation of the dependency-based compositional method so as to process sentences with open syntax. As one of the main limitations of the current DepFunc version is that it cannot be applied to syntactically unrestricted expressions, we are working to improve the method so that it can be more general and no dependent of specific syntactic constructions.

%As it is a method motivated by the linguistic structure, the system should be improved by considering the linguistic knowledge of specialists. Future innovations may come from incorporating key concepts from Cognitive Grammar \citep{Langacker1987}, as mentioned earlier, and from Construction Grammar \citep{Goldberg1995}. Concerning the latter, the proposed compositional method might be improved with usage-based syntactic-semantic constructions. A syntactic-semantic construction would be defined as a sequence of paradigmatic classes represented as slot-constraints at different levels of generalization, from syntactic to lexical level.  
 %In this way, for instance, it would be possible to define the syntactic class of nouns playing the role of subject in a bi-transitive construction without specifying the lexical units of any constituent. This gives rise to a contextualized vector representing the very generic meaning of the subject position in a bi-transitive construction. On the other hand, and in a complementary way, our approach allows us to define fully lexicalized paradigms. Such a semantic flexibility will be considered in further implementations of the system.

On the other hand, special attention will be paid to the compositional meaning of referential and grammatical expressions, namely determiners, auxiliary verbs or deictic adverbs, whose combination with other lexical words is not driven by selectional preferences but by mechanisms involving grounding strategies to the discourse situation \citep{Gupta2015}. It should be noted that our approach (as well as the attention mechanism) only addresses the local lexical meaning, leaving out the global one referring to knowledge about situations, circumstances and state of affairs \citep{Erk&Herbelot2021}. So, a crucial challenge we will address is to find a way to represent the global meaning so that it interacts with the local/lexical one.

We will also address challenges related to the degree of compositionality, which includes the treatment of expressions such as idioms, compounds, or collocations. It is likely that we need other non-compositional or weakly compositional semantic mechanisms, other than that based on selection preferences, to account for this type of expressions. 

Another possible improvement to be explored is to propose models that take into account both types of relations, syntagmatic and paradigmatic, in order to overcome the syntactic limitations of both the attention mechanism and our approach based on selectional preferences.

Finally, we will also explore the possibility of elaborating neuro-symbolic strategies by integrating explicit linguistic information into the attention mechanism of Transformers. This symbolic information could distill the neural architecture of large language models, by guiding the readjustment of weights without resorting to the brute force of an uninterpretable black box,  as most of these giant Transformers operate today.

DepFunc software and the evaluation datasets used in the multilingual experiments are freely available.\footnote{\url{https://github.com/gamallo/DepFunc}}

%TC:ignore
\begin{acknowledgement}
 This research was funded by: project "Nós - Galician in the society and economy of artificial intelligence", agreement between Xunta de Galicia and University of Santiago de Compostela; Horizon Europe, Marie Skłodowska-Curie Actions (MSCA), Doctoral Networks; European Union; LingUMT, grant PID2021-128811OA-I00, MEC; DeepR, grant TED2021-130295B-C31, MEC; Big-eRisk, grant PLEC2021-007662, MEC; and grant ED431G2019/04 by the Galician Ministry of Education, University and Professional Training, and the European Regional Development Fund (ERDF/FEDER program), and Groups of Reference: ED431C 2020/21.
\end{acknowledgement}
%TC:endignore

\bibliographystyle{apalike}
\bibliography{article}
\end{document}